\documentclass[conference]{IEEEtran}
\IEEEoverridecommandlockouts
\usepackage{cite}
\usepackage{amsmath,amssymb,amsfonts}
\usepackage{graphicx}
\usepackage{caption}
\usepackage{textcomp}
\usepackage{xcolor}
\usepackage{algorithm}
\usepackage{algpseudocode}
\usepackage{multirow}
\def\BibTeX{{\rm B\kern-.05em{\sc i\kern-.025em b}\kern-.08em
    T\kern-.1667em\lower.7ex\hbox{E}\kern-.125emX}}
\begin{document}

\title{
Physics Informed Recurrent Neural Networks for Seismic Response Evaluation of Nonlinear Systems\\
}

\author{
\IEEEauthorblockN{Faisal Nissar Malik}
\IEEEauthorblockA{\textit{Dept. of Civil Engineering} \\
\textit{Lehigh University}}
\and
\IEEEauthorblockN{James Ricles}
\IEEEauthorblockA{\textit{Dept. of Civil Engineering} \\
\textit{Lehigh University}}
\and
\IEEEauthorblockN{Masoud Yari}
\IEEEauthorblockA{\textit{Dept. of Computer Science} \\
\textit{Lehigh University}}
\and
\IEEEauthorblockN{Malik Arsala Nissar}
\IEEEauthorblockA{\textit{Dept. of Mechanical Engineering} \\
\textit{IIT Kanpur}}
}

\maketitle

\section{Abstract}
Dynamic response evaluation in structural engineering is the process of determining the response of a structure, such as member forces, node displacements, etc when subjected to dynamic loads such as earthquakes, wind, or impact. This is an important aspect of structural analysis, as it enables engineers to assess structural performance under extreme loading conditions and make informed decisions about the design and safety of the structure. Conventional methods for dynamic response evaluation involve numerical simulations using finite element analysis (FEA), where the structure is modeled using finite elements and the equations of motion are solved numerically. Although effective, this approach can be computationally intensive and may not be suitable for real-time applications. To address these limitations, recent advancements in machine learning, specifically artificial neural networks, have been applied to dynamic response evaluation in structural engineering. These techniques leverage large data sets and sophisticated algorithms to learn the complex relationship between inputs and outputs, making them ideal for such problems.
In this paper, a novel approach is proposed for evaluating the dynamic response of multi-degree-of-freedom (MDOF) systems using physics-informed recurrent neural networks.
The focus of this paper is to evaluate the seismic (earthquake) response of nonlinear structures. The predicted response will be compared to state-of-the-art methods such as FEA to assess the efficacy of the physics-informed RNN model.

\section{Introduction}
Dynamic response analysis is a valuable tool for designing and assessing the performance of structures under a variety of loads such as seismic, wind, or wave loading, as well as for conducting reliability analyses of infrastructure and large urban areas. 
The traditional approach for dynamic response analysis typically involves creating numerical models and solving the partial differential equations using numerical integrators, such as the Newmark-$\beta$ method or the KR-$\alpha$\cite{Kolay2014} method to predict the system's response under dynamic loads. 
However, this approach is computationally expensive and is not feasible in scenarios such as probabilistic seismic analysis where a large number of simulations need to be performed,  or real-time hybrid simulations involving structures with a large number of degrees of freedom where the computations need to be performed in real-time.\\
Machine learning and Artificial intelligence have proven to be effective tools for predicting the response of complex nonlinear systems with a high degree of accuracy at a fraction of the computational cost compared to traditional finite element analysis.
Furthermore, machine learning techniques can also be leveraged as surrogate low-fidelity models to predict the response of systems under insufficient prior knowledge of the systems.
These techniques can also be leveraged to model the system's behavior based on experimental data.\\
The use of machine learning in predicting the behavior of structures under dynamic loads has been investigated by researchers such as Lagaros\cite{Lag2012}, Zhang \cite{Zhang2019a,Zhang2020}, and Eshkevari\cite{ssesh2020}. 
These studies demonstrate that machine learning has a lot of potential in the area of structural dynamics and response predictions. 
Zhang\cite{Zhang2020} used physics based LSTM models (PhyLSTM\textsuperscript{2}, PhyLSTM\textsuperscript{3}) for seismic response prediction.
Their proposed architecture included two or three separate LSTM-based neural network models to predict the system's response such as the displacement and the restoring force. 
Essentially, they added additional terms in the loss function that penalizes the neural network based on discrepancies in the equations of motion.
Even though the accuracy of the predictions was enhanced by imposing these new physical constraints, these architectures lack guidance from physics resulting in overly complex networks that require extended training periods and large amounts of training data. 
Therefore, in this paper, a new physics-guided neural network architecture is proposed that uses the idea of numerical integrators to predict the seismic response of highly nonlinear structural systems under limited availability of training data. 
The network architecture of the model is described in detail in Section \ref{label:inference}.

\section{Network Architecture}
\label{label:inference}
\begin{figure}
    \centering
    \includegraphics[width=0.45\textwidth]{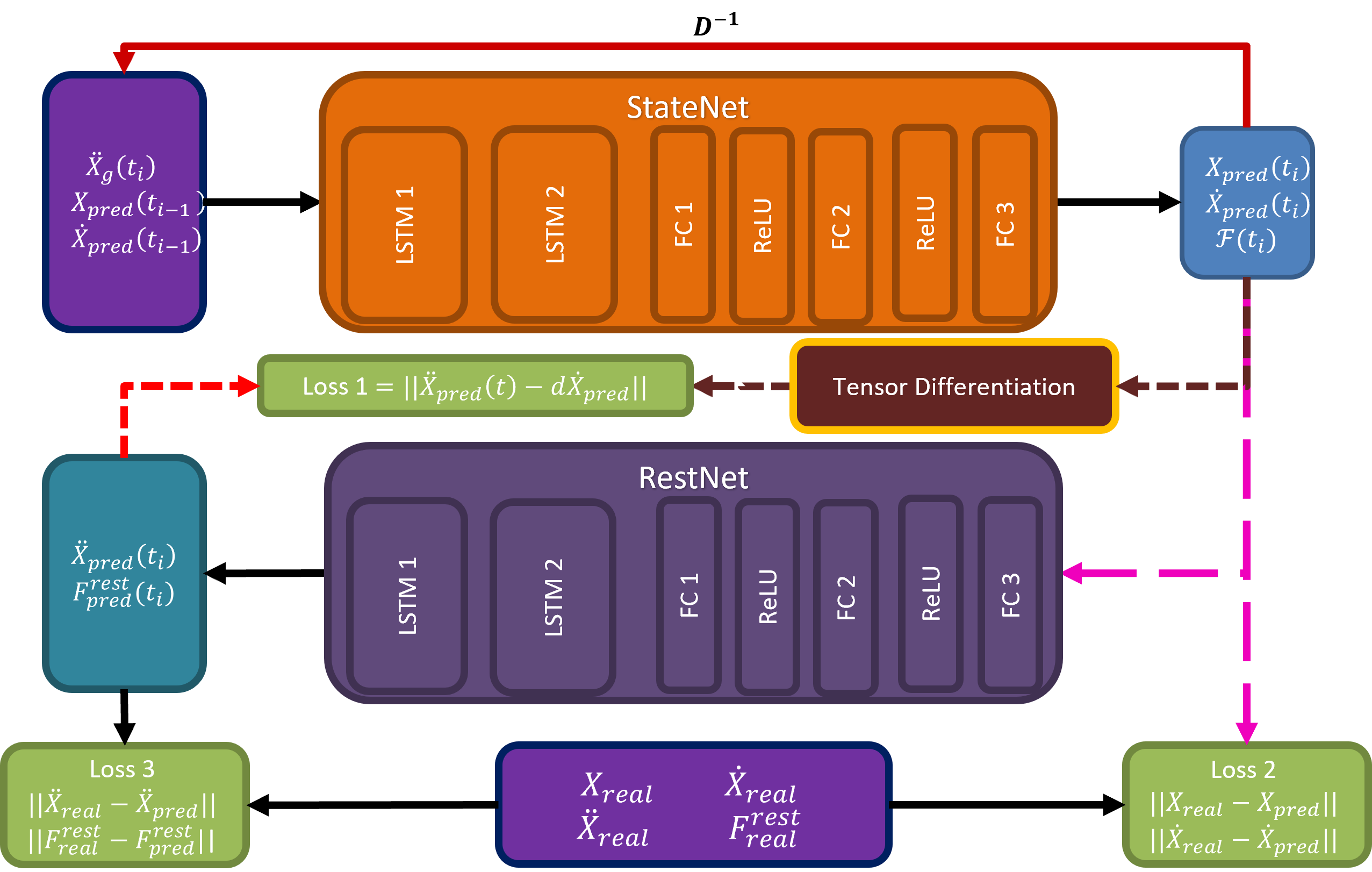}
    \caption{Architecture of the proposed model}
    \label{fig:Model_architecture}
\end{figure}

Numerical solvers determine the structure's response $X=\big\{x(t), \dot{x}(t), \ddot{x}(t)\big\}$, where $x(t)$ is the displacement, $\dot{x}(t)$ is the velocity and $\ddot{x}(t)$) is the acceleration at time $t$ from the state of the system $\big\{x(t_{i-1}), \dot{x}(t_{i-1}), \ddot{x}(t_{i-1})\big\}$ at time $t_{i-1}$ and the forcing function $f(t_i)$ at time $t$. For example the KR-$\alpha$ method \cite{Kolay2014} gives:
\begin{align}
    \dot{X}_{i+1}=\dot{X}_{i}+\Delta t \alpha_1 \ddot{X}_i\\
     X_{i+1}=X_{i}+\Delta t \dot{X}_{i+1}+\Delta t^2 \alpha_2 \ddot{X}_i\\
     M\hat{\ddot{X}}_{i+1}+C \dot{X}_{i+1-\alpha_f}+K X_{i+1-\alpha_f}=F_{i+1-\alpha_f}\\
\end{align}
where,
\begin{align}
        \hat{\ddot{X}}_{i+1} & = (I-\alpha_3)\ddot{X}_{i+1}+\alpha_3\ddot{X}_{i}\\
        \dot{X}_{i+1-\alpha_f} & = (1-\alpha_f)\dot{X}_{i+1}+\alpha_f\dot{X}_{i}\\
        X_{i+1-\alpha_f} & = (1-\alpha_f)X_{i+1}+\alpha_f X_{i}\\
        F_{i+1-\alpha_f} & = (1-\alpha_f)F_{i+1}+\alpha_f F_{i}
\end{align}
and initial acceleration vector is given by $M\ddot{X}_0=F_0-C\dot{X}_0-KX_0$, where
$M,C,$ and $K$ are the mass, damping, and stiffness matrices,
respectively, of dimension $n \times n$ for an MDOF system with $n$ DOFs; 
$X \dot{X}, \ddot{X},$ and $F$ are the displacement, velocity, acceleration, and force vector, respectively;
$\alpha_1, \alpha_2, \alpha_3$ are the integration parameter matrices of dimensions $n\times n$; 
I is the identity matrix of dimension $n \times n$; and $\alpha_f$  is a scalar integration parameter.\\
The network architecture is based on the same principle and is shown in Figure \ref{fig:Model_architecture}.
The input to the model at time $t$ is the state of the system [$x(t_{i-1})$, $\dot{x}(t_{i-1})$ 
] at time $t_{i-1}$ and the ground acceleration \big\{$\ddot{x_g}(t)$\big\} at time $t$, and the model predicts the system response at time $t$.
The model consists of two separate subnets, the state prediction model (StateNet) and the restoring force and acceleration prediction model (RestNet). 
The StateNet takes the state of the system \big\{$x(t_{i-1})$, $\dot{x}(t_{i-1})$
\big\} and the ground acceleration \big\{$\ddot{x_g}(t)$\big\} to calculate the state of the system at time $t$. The RestNet takes the state of the system predicted by the StateNet at time $t$ to predict the acceleration and restoring force at time $t$.
The two subnets consists of two LSTM layers, followed by three fully connected layers as shown in the figure. 
Each LSTM layer has 200 neurons and the output from LSTM layers is passed through the three fully connected dense layers and ReLU activation to get the final output.
The model architecture is shown in Figure \ref{fig:Model_architecture}.\\
The outputs of the model are passed through a tensor differentiator to calculate the time derivatives of the responses. The loss of model is calculated from these time derivatives and the predicted responses as:
\begin{equation}
    L(\theta_1,\theta_2)= w_1 L_1(\theta_1)+w_2 L_2(\theta_1)+w_3 L_3(\theta_1,\theta_2)
\end{equation}
where $\theta_1$, and $\theta_2$ are the parameters of the state and acceleration prediction models, respectively, $L(\theta_i)$ are the loss functions, $w_i$ is the weight given to $i^{th}$ loss term, and the loss terms are:
\begin{align}
    L_1(\theta_1) &= ||\dfrac{dX_{pred}}{dt}-\dot{X}_{pred}||\\
    L_2(\theta_1) &= ||X_{real}-X_{pred}||+||\dot{X}_{real}-\dot{X}_{pred}||\\
    L_3(\theta_1,\theta_2) &= ||\ddot{X}_{real}-\ddot{X}_{pred}||\\
\end{align}
$L_1(\theta_1)$ forces the time derivative of predicted response to be equal to the predicted derivative of the response.
$L_2(\theta_2)$ and $L_3(\theta_1,\theta_2)$ are mean square losses for the two subnets.
For simplicity, the values of $w_i's$ are taken to be unity.
It should be observed that $L_3$ depends on both $\theta_1$ and $\theta_2$, therefore, the problem is a bi-objective optimization problem.\\
The network is trained by using two different approaches. 
First, the model is trained for $N$ epochs with teacher forcing. 
In this approach, the targets $x_{real}(t_{i-1})$ and $\dot{x}_{real}(t_{i-1})$ are passed as inputs to the StateNet instead of the model's predictions at time $t_{i-1}$ to get the state at time $t$ and the outputs from the StateNet are passed as inputs to the RestNet to get the acceleration and restoring forces.
The Algorithm for teacher forcing is shown in Algorithm \ref{alg:capTF}.\\
\begin{algorithm}[H]
\caption{Training the Model with Teacher Forcing for One Input Sequence}\label{alg:capTF}
\begin{algorithmic}[1]
\State $X_0 \gets 0$, $\dot{X}_0 \gets 0$, $\ddot{X}_0 \gets F_0/M$
\While{$t_i \le t_n$}
\State $\small{X_{p}(t_i)} \gets \small{\text{Subnet}^1(X_{r}(t_{i-1}),\dot{X}_{r}(t_{i-1}), \ddot{X}_{r}(t_{i-1})})$
\State $\small{\dot{X}_{p}(t_i)} \gets \small{\text{Subnet}^1(X_{r}(t_{i-1}),\dot{X}{r}(t_{i-1}), \ddot{X}_{r}(t_{i-1}))}$
\State $\small{\ddot{X}_{p}(t_i)} \gets \small{\text{Subnet}^2(X_{r}(t),\dot{X}_{r}(t), \ddot{X}_g(t))}$
\EndWhile
\State $L(\theta_1,\theta_2) \gets L_1(\theta_1)+L_2(\theta_2)+L_3(\theta_1,\theta_2)$
\State $\theta_i \gets \theta_i+lr\dfrac{dL(\theta_1,\theta_2)}{d\theta_i}$
\end{algorithmic}
\end{algorithm}
After training the model with teacher forcing, the model is then trained for $M$ epochs with scheduled learning. 
In this approach, the model is trained as shown in the earlier algorithm, however, the training data is randomly replaced by the model's own predictions to help the model correct its own mistakes.
The trained model is used to predict the response to unknown ground motions as described below.\\
To predict the response to an unknown ground motion, the model is called one step at a time to predict the responses at time $t_i$.
The predicted response through feedback is passed back into the model to calculate the response at time $t_{i+1}$. 
The displacement, velocity, and acceleration are initialized as $X_0$, $\dot{X}_0$, and $\ddot{X}_0=(F_0-C\dot{X}_0-KX_0)/M$, respectively.
Since the system considered in this study starts from at rest conditions i.e., $X_0=\dot{X}_0=0$, the initial acceleration $\ddot{X}_0$ becomes $\ddot{X}_0=F_0/M=\Gamma\ddot{X}_g(0)$, where $\Gamma$ is the influence vector.
The pseudocode for the model inference is shown in Algorithm \ref{alg:capTFN}.
\begin{algorithm}[H]
    \caption{Model Inference}\label{alg:capTFN}
\begin{algorithmic}[1]
\State $X_0 \gets 0$, $\dot{X}_0 \gets 0$, $\ddot{X}_0 \gets F_0/M$
\While{$t_i \le t_n$}
\State $\small{X_{p}(t_i)} \gets \small{\text{Subnet}^1(X_{p}(t_{i-1}),\dot{X}_{p}(t_{i-1}), \ddot{X}_{p}(t_{i-1})})$
\State $\small{\dot{X}_{p}(t_i)} \gets \small{\text{Subnet}^1(X_{p}(t_{i-1}),\dot{X}{p}(t_{i-1}), \ddot{X}_{p}(t_{i-1}))}$
\State $\small{\ddot{X}_{p}(t_i)} \gets \small{\text{Subnet}^2(X_{p}(t),\dot{X}_{p}(t), \ddot{X}_g(t))}$
\EndWhile
\end{algorithmic}
\end{algorithm}
In the proposed architecture, the learning rate of the optimizer, the number of neurons in the layers, the number of connections between the StateNet and RestNet, and the batch size are the most important hyper-parameters. 
In this architecture, no regularization methods have been used.
Note that the hyper-parameters are problem specific and need to be fine-tuned as per the problem. 
The model is trained on a relatively small batch size. 
Since, the training involves tensor differentiation, and the time series are large in length ($\geq1500$), selecting a high batch size is not possible due to memory constraints.
A batch size of five was selected for training the model. 
At this batch size, the model was observed to perform the best.\\
The data-driven model used for benchmarking the performance of the proposed network is a LSTM-based network developed by Zhang \cite{Zhang2019a}.
The input to the data-driven model is the ground acceleration and the model predicts the acceleration, velocity, displacement, and the restoring force. 
The data-driven model is trained using Adam optimizer with MSE as the loss function.
\section{Numerical examples: Dataset}
\begin{figure}
    \centering
    \includegraphics[width=0.45\textwidth]{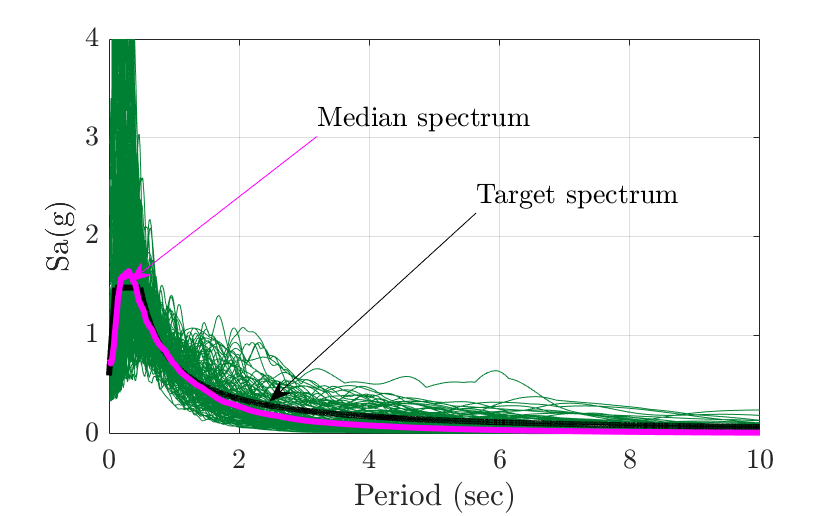}
    \caption{The response spectrum of the ground motion set used for the second numerical example.}
    \label{fig:spectrum}
\end{figure}

\begin{figure}
    \centering
    \includegraphics[width=0.45\textwidth]{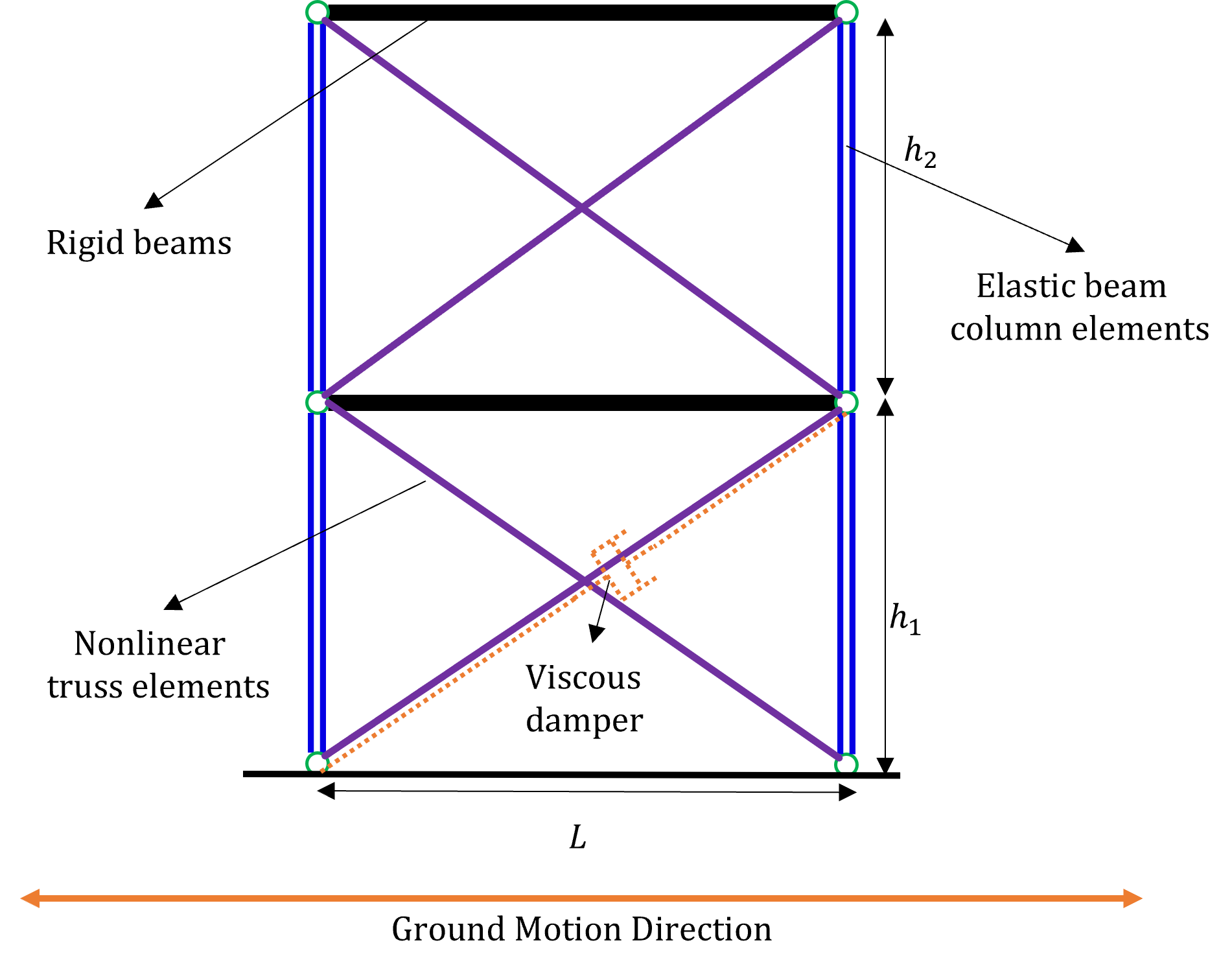}
    \caption{Two-story nonlinear damped building frame considered for the study}
    \label{fig:building}
\end{figure}

The data used for training the model is generated using finite element analysis.
However, experimental  or a combination of experimental and simulated data can also be used to train the model. 
To generate the training data, a set of $D$ ground motions is selected and scaled to match a target design spectrum.
The structure's response for the ground motions is calculated through finite element analysis and numerical integration to generate the training and validation dataset.\\
For this paper, two numerical examples are investigated. 
The first example is a single degree of freedom (SDOF) Boucwen system. 
This system and dataset are the same as those used by Zhang\cite{Zhang2020}. 
The primary purpose of using this dataset was to benchmark the proposed architecture's performance against the PhyLSTM\textsuperscript{3} model developed by Zhang\cite{Zhang2020}.
The dataset used by Zhang\cite{Zhang2020} consisted of 85 synthetic ground motions. 
The ground motion duration is $30$\,second and the ground motions are sampled at a frequency of $50$\,Hz. Therefore, each ground motion has 1501 data points.
60 examples were used for training the model and 25 examples were used for evaluating the model.
The model is compared to the data driven model \cite{Zhang2019a}, and the PhyLSTM\textsuperscript{3} model\cite{Zhang2020}.\\
For the second example, a nonlinear MDOF system as shown in Figure \ref{fig:building} is used. 
The frame has a mass of $6,720$\,kN, and $10,080$\,kN for the first and second floors, respectively. 
The beams are rigid, and the columns are modeled with elastic beam-column elements.
The bracing is modeled using a nonlinear truss element with "Steel02" uniaxial material.
The material has a post-yield stiffness ratio of 2.8\%.
The ground floor is also connected to nonlinear viscous dampers with a velocity exponent of 0.2, as shown in the figure. 
The building is modeled in OpenSees and a combination of real world ground motion records downloaded for PEER NGA west 2 database \cite{ancheta2013peer} and synthetic ground motions is used to generate the training and validation dataset.
Numerical integration is performed using the HHT-$\alpha$ method to obtain the structure's response.\\
The dataset consists of 274 ground motions sampled at a frequency of $50$\,hz, and the ground motion duration is fixed at 30\,sec. 
The sampled records are scaled to match the UHS MCE design response spectrum in the period range $0.4$\,sec to $2.0$\, sec i.e., $0.5\times T_1$ to $2.5\times T_1$, where $T_1$ is the fundamental period of the structure. 
The wavelet record scaling procedure proposed by Montejo\cite{montejo2021response} is used for scaling the ground motions to the target spectrum
The design response spectrum, median response spectrum of the ground motion set and the individual scaled records are shown in Figure \ref{fig:spectrum}. 
As can be seen in the figure, the response spectrum matches well with the target spectrum in the period range of interest.\\
The output from the network is the displacement, velocity, and acceleration response of the two floors, the force in damper at first floor, and the force in the first floor brace. 
130 examples are used for training the model and 144 are used to validate the model.
The model is trained as explained earlier in Section \ref{label:inference}. The purely data-driven LSTM model described earlier in Section \ref{label:inference} is also trained and evaluated as a benchmark.
\section{Numerical Examples: Results}
In this section, the results for the two numerical examples discussed in the earlier section are presented below.
\subsection{Numerical Example I: SDOF Boucwen System}
\begin{figure}
    \centering
    \includegraphics[width=0.48\textwidth]{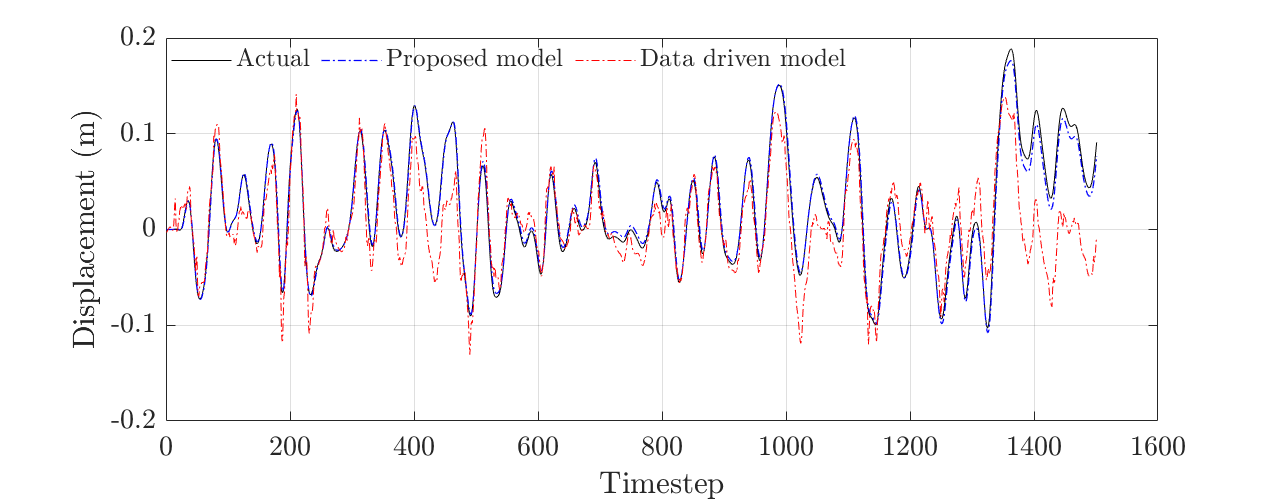}
    \includegraphics[width=0.48\textwidth]{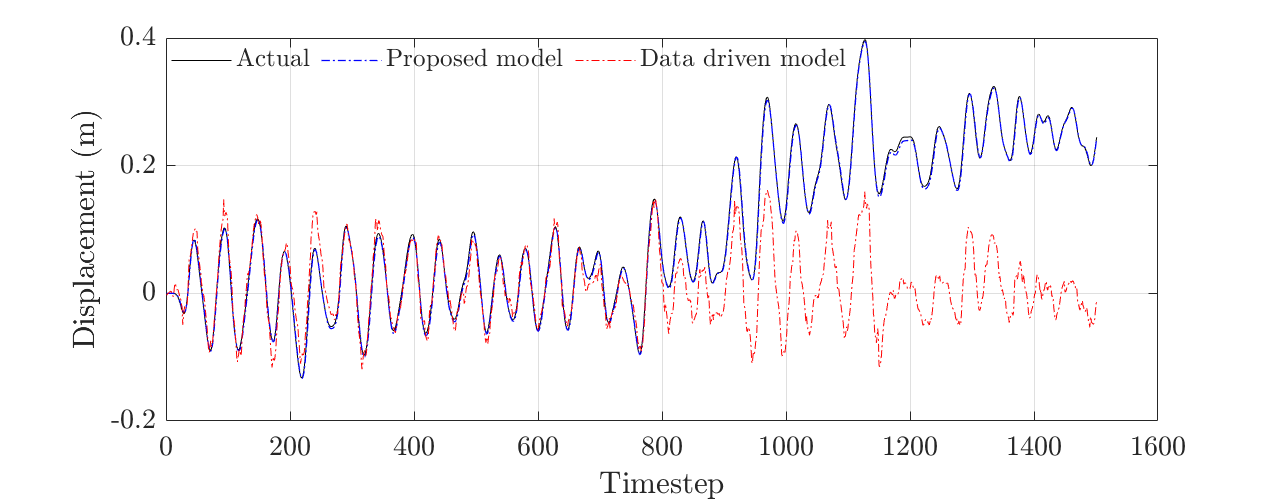}
    \caption{The actual and predicted response for two examples in the validation dataset.}
    \label{fig:dd_bw_resp}
\end{figure}

\begin{figure}
    \centering
    \includegraphics[width=0.23\textwidth]{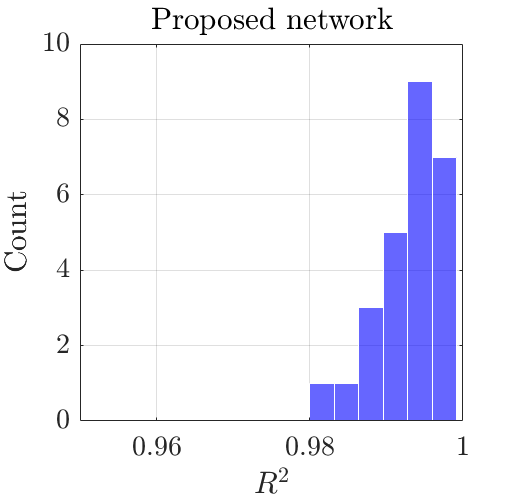}
    \includegraphics[width=0.23\textwidth]{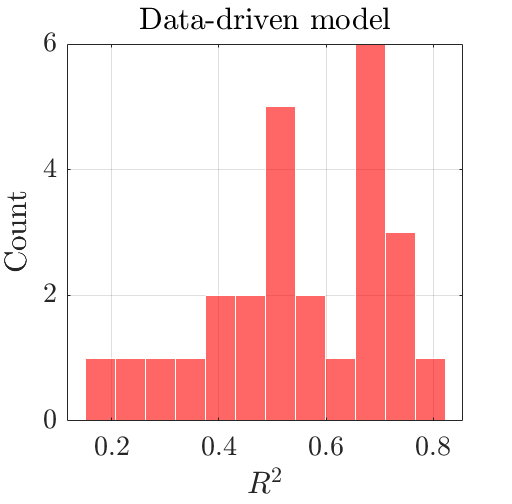}
    \caption{Histogram of $R^2$ values of the displacement response predicted by the (a) proposed model and (b) data-driven model on the validation data set.}
    \label{fig:R2_bW}
\end{figure}

In this section, the results of the SDOF Boucwen system are presented. 
Recall that the training data consisted of 85 examples, where-in 60 examples were used for training the model and 25 examples were used for evaluating the model. 
The data-driven model is trained for 200 epochs and the training data is shuffled before each epoch. 
The proposed architecture is trained for 100 epochs with teacher forcing and 100 epochs with scheduled learning as described earlier in Section \ref{label:inference}. \\
Figure \ref{fig:dd_bw_resp} shows the actual displacement response of the structure, the response predicted using the proposed architecture, and the response predicted using the data-driven model on two ground motions from the validation data set. 
As can be seen in the figure, the proposed model is much better at predicting the displacement response of the structure in comparison to the purely data-driven LSTM model. 
The $R^2$ values for the validation dataset are calculated as:
\begin{equation}
    R^2(i)=1-\dfrac{SS_{res}(i)}{SS_{tot}(i)}
\end{equation}
where, $R^2(i)$ is the R-square value on i\textsuperscript{th} example, $SS_{res}(i)$ is the sum of squares of the residuals for  i\textsuperscript{th} example and $SS_{tot}(i)$ is the total sum of squares for i\textsuperscript{th} example. 
The $R^2$ values for the proposed architecture and the purely data-driven model for the 25 validation examples are calculated and are shown in Figure \ref{fig:R2_bW}.
The mean of $R^2$ is $0.9929$, and the standard deviation of $R^2$ is $0.0043$ for the proposed architecture on the validation dataset. 
These values are $0.5460$, and $0.1660$ for the data-driven model.
In other words, the proposed model is able to explain $99.29$\% variation in displacement response, while the data-driven model is able to explain only $54.60$\% variation in the displacement response.\\
Since the system under consideration and the dataset used for this example is the same as used by Zhang\cite{Zhang2020}, the results are compared to the PhyLSTM\textsuperscript{3} model proposed by them.
The code for PhyLSTM\textsuperscript{3} is not publicly available and thus it was not possible to reproduce the results. 
Therefore, the data shown by them in their original paper was compared to the proposed architecture.
The $R^2$ values for the best and worst case of the entire dataset for PhyLSTM\textsuperscript{3} are $0.99$ and $0.79$, respectively\cite{Zhang2020}. 
The $R^2$ values for the best and the worst case on the entire dataset for the proposed architecture are $0.995$ and $0.981$, respectively. 
Thus it can be seen that the proposed architecture outperforms the PhyLSTM\textsuperscript{3} model for this system and dataset.\\
It is also noteworthy to look at the mean square error of the predicted outputs. 
The mean square error for the data-driven LSTM and the proposed architecture are summarized in table \ref{tab:r2_bw}.
\begin{table}[h]
    \centering
    \begin{tabular}{|c|c|c|}
        \hline
         Response & Proposed architecture & Data-driven model \\ \hline
         Displacement & $9.14 \times 10^{-5}$ & $6.1 \times 10^{-3}$ \\ \hline
         Velocity & $2.33 \times 10^{-5}$ & $2.12 \times 10^{-4}$\\ \hline
         Acceleration & $1.1 \times 10^{-3}$ & $2.8 \times 10^{-3}$ \\ \hline
    \end{tabular}
    \caption{The mean square error for the proposed architecture and the data-driven model on the validation dataset.}
    \label{tab:r2_bw}
\end{table}
As can be seen in the table, the proposed model shows a reduction of $98.50$\% in the mean square error for displacement response and a reduction of $89$\% in the mean square error for velocity response on the validation dataset when compared to the purely data-driven LSTM model.
\subsection{Numerical Example II: Nonlinear MDOF building.}
\begin{figure}
    \centering
    \includegraphics[width=0.48\textwidth]{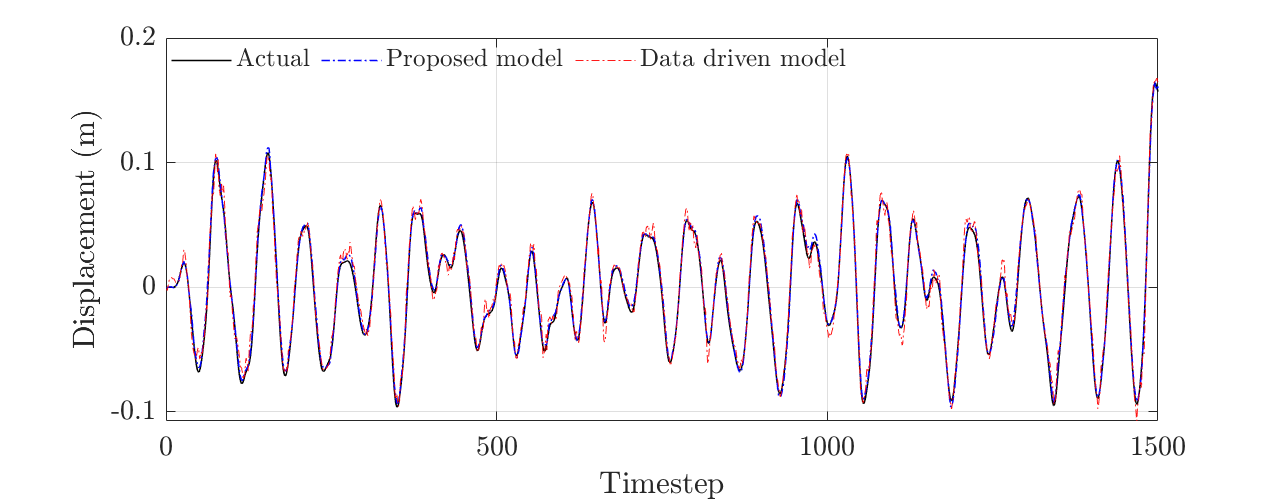}
    \includegraphics[width=0.48\textwidth]{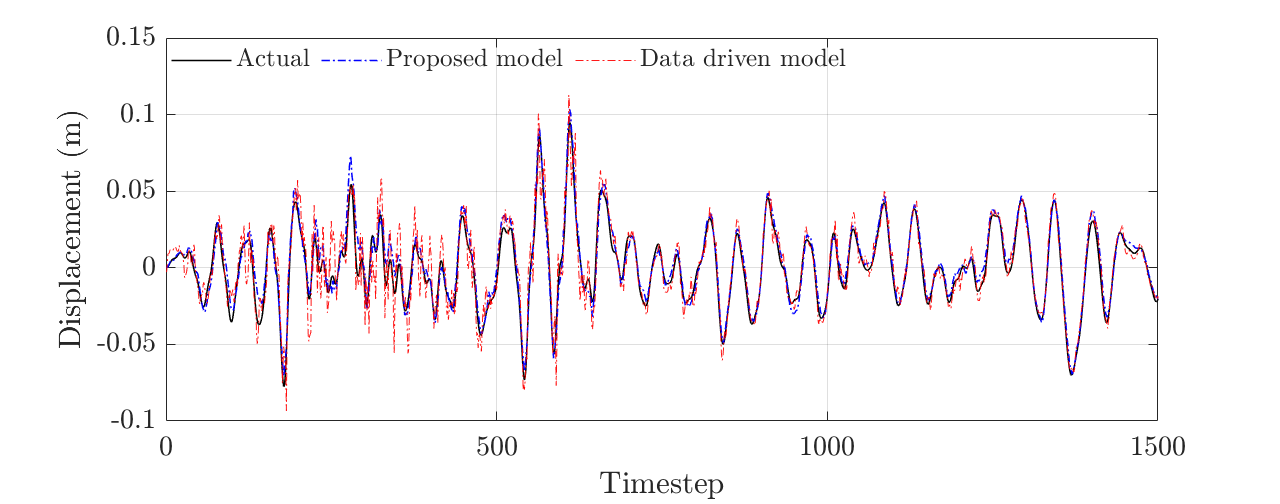}
    \caption{The actual and predicted displacement response of two examples in the validation dataset for the MDOF frame system.}
    \label{fig:dd_bf_resp}
\end{figure}

\begin{figure}
    \centering
    \includegraphics[width=0.23\textwidth]{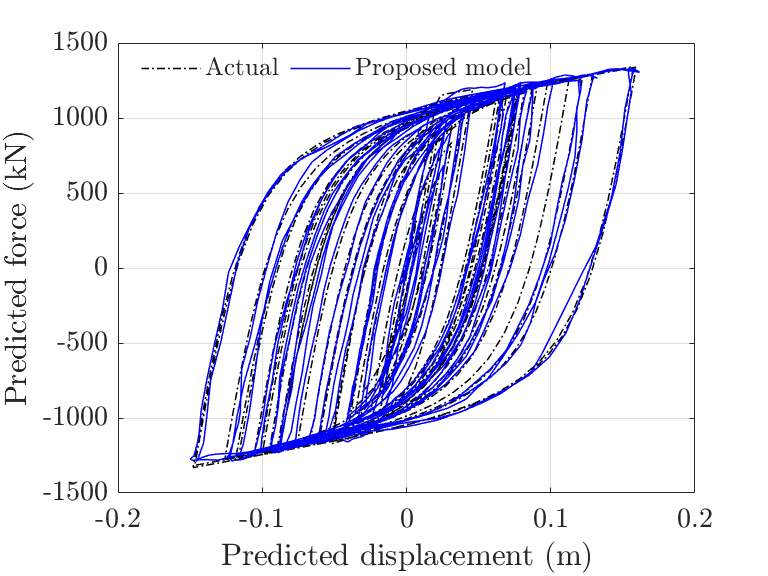}
    \includegraphics[width=0.23\textwidth]{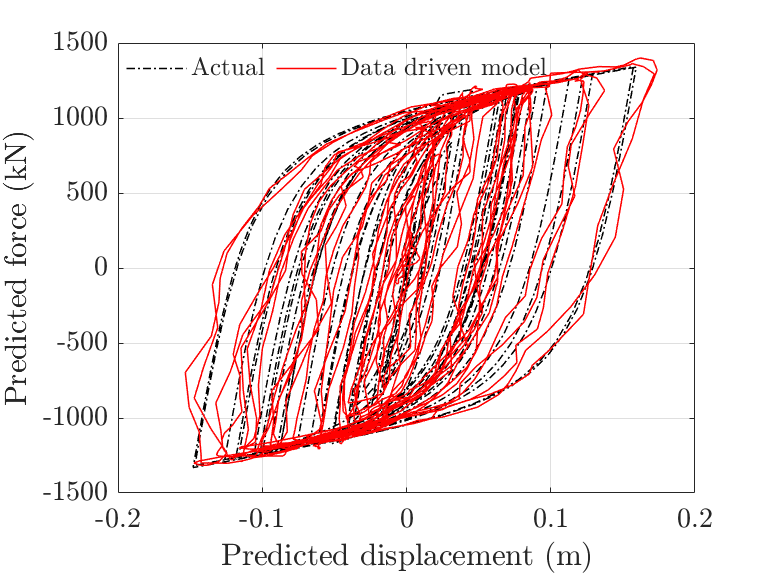}
    \caption{Predicted force vs predicted deformation behavior of the first-floor brace. The first plot shows the results from the proposed network and the second plot shows the results from the data-driven model.}
    \label{fig:hyst}
\end{figure}

\begin{figure}
    \centering
    \includegraphics[width=0.23\textwidth]{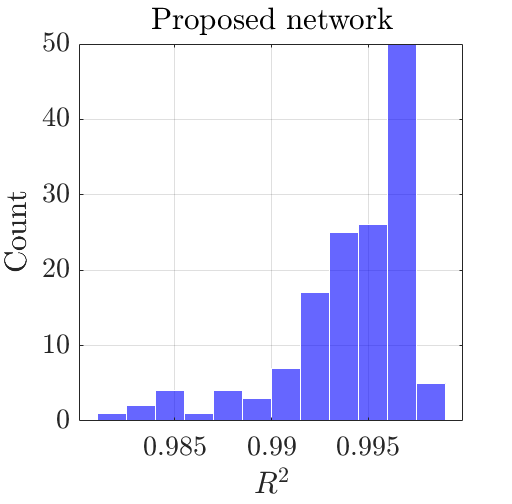}
    \includegraphics[width=0.23\textwidth]{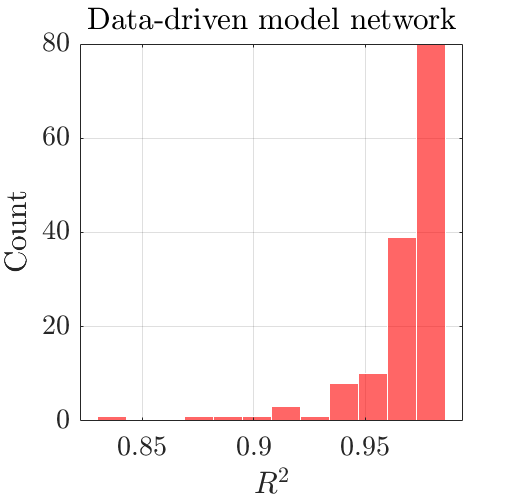}
    \caption{Histogram of the $R^2$ values of the predicted displacement response from (a) the proposed network and (b) the data-driven model on the validation dataset.}
    \label{fig:r2_mdof}
\end{figure}

\begin{figure}[h]
    \centering
    \includegraphics[width=0.35\textwidth]{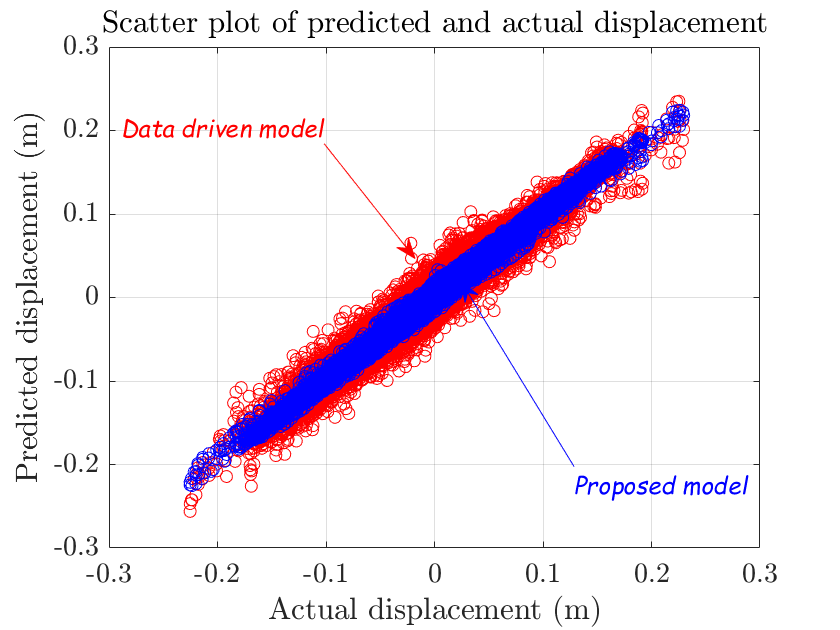}
    \caption{Scatter plot of the predicted and actual displacement.}
    \label{fig:scatter}
\end{figure}

In this section, the results for the nonlinear building frame are presented. 
Recall that the training data consisted of 274 examples, where-in 130 examples are used for training the model and 144 examples are used for evaluating the model. 
The data-driven model is trained for 500 epochs and the training data is shuffled before each epoch. 
The proposed architecture is trained for 250 epochs with teacher forcing and 250 epochs with scheduled learning as described earlier in Section \ref{label:inference}.\\
Figure \ref{fig:dd_bf_resp} shows the predicted displacement response for the first floor for two ground motions in the validation dataset. 
As can be seen in the figure, the proposed architecture is able to predict the response of the structure with a high degree of accuracy. 
The figure also shows that the output from the data-driven model is noisier in comparison to the response from the proposed architecture.
Furthermore, the data-driven model overestimates the predicted displacement response.
Figure \ref{fig:hyst} shows the predicted force vs deformation (hysteretic) behavior of the first-floor brace of the proposed architecture and the data-driven model for one ground motion record in the validation dataset.
As can be seen in the figure, the proposed architecture is much better at capturing the nonlinear hysteretic behavior in comparison to the purely data-driven model.\\ 
The $R^2$ and mean square error values of the predicted responses from the proposed architecture and purely data-driven LSTM model are also calculated and are shown in table \ref{tab:r2}.
As can be seen in the table, the proposed architecture outperforms the purely data-driven model for all the response quantities of interest, thus highlighting the efficacy of the proposed architecture. 
The mean square error of the predicted responses from the proposed network is roughly one order of magnitude less than that of the same from the data-driven model.
For illustration, a histogram of $R^2$ values of the displacement response on the validation data set from the proposed network and the data-driven model is shown in figure \ref{fig:r2_mdof}. 
The figure shows that the $R^2$ values of the predicted responses from the proposed network are greater than $0.98$, indicating a very strong correlation between the predicted and actual response.
On the other hand, the $R^2$ values of the predicted responses from the purely data-driven model lie anywhere between $0.8$ and $0.98$. 
The picture is even more clear when we look at the scatter plot of the actual and predicted displacement for the proposed architecture and the data-driven model shown in figure \ref{fig:scatter}. 
As can be seen in the figure, the scatter plot of the actual displacement and the output from the proposed model is much more correlated than the same from the purely data-driven model. Since this example was not highly nonlinear, the data-driven model was also able to predict the response of the system with good accuracy.
\begin{table}
    \centering
    \begin{tabular}{|c|c|c|c|c|}
        \hline
         \multirow{2}{*}{Response} & \multicolumn{2}{c|}{Proposed architecture}  & \multicolumn{2}{c|}{Data driven model} \\
          & $R^2$ & MSE & $R^2$ & MSE \\ \hline 
         Displacement & $0.9935$ & $3.14 \times 10^{-7}$ & $0.9679$ & $3.09 \times 10^{-6}$ \\ \hline
         Velocity & $0.9957$ & $6.34 \times 10^{-7}$ & $0.9843$ & $1.21 \times 10^{-5}$ \\ \hline
         Acceleration & $0.9949$ & $7.1 \times 10^{-3}$ & $0.9962$ & $8.3 \times 10^{-3}$ \\ \hline
         Force & $0.9898$ & $1.1 \times 10^{-5} $ & $0.9897$ & $2.2165 \times 10^{-4}$\\ \hline
    \end{tabular}
    \caption{$R^2$ and mean square error on the validation dataset for the proposed model and data-driven model.}
    \label{tab:r2}
\end{table}

\section{Summary and Conclusions}
In this paper, a new architecture is proposed for evaluating the seismic response of nonlinear structures. 
The proposed algorithm was able to capture the nonlinear response of the structures with a high degree of accuracy.
Based on the results of the study, the following conclusions have been made:
\begin{enumerate}
    \item The proposed architecture is able to capture the nonlinear behavior of structures accurately with less training data requirement.
    \item Combined with teacher forcing and scheduled learning, the proposed architecture doesn't require long training epochs and performs well without the need for data scaling or normalization.
    \item The proposed architecture is able to capture the hysteretic behavior of structures with great accuracy.
    \item The proposed architecture is able to predict the seismic response of the structure in a fraction of the time compared to finite element analysis and thus can be used in combination with  real-time hybrid simulations or  probabilistic seismic analysis where computation cost is a huge concern.
\end{enumerate}

\bibliographystyle{plain}
\bibliography{mybibfile}

\end{document}